\PassOptionsToPackage{prologue,dvipsnames}{xcolor}
\documentclass[sigconf, nonacm]{acmart}

\newcommand\vldbdoi{XX.XX/XXX.XX}
\newcommand\vldbpages{XXX-XXX}
\newcommand\vldbvolume{14}
\newcommand\vldbissue{1}
\newcommand\vldbyear{2025}
\newcommand\vldbauthors{\authors}
\newcommand\vldbtitle{\shorttitle} 
\newcommand\vldbavailabilityurl{URL_TO_YOUR_ARTIFACTS}
\newcommand\vldbpagestyle{plain} 

\usepackage{amssymb}
\usepackage{amsmath} 
\usepackage{booktabs} 
\usepackage{array}
\usepackage{graphicx} 
\usepackage{multirow}
\usepackage{color}
\usepackage{makecell} 
\usepackage{float}
\usepackage{subcaption}
\usepackage{colortbl}
\usepackage{placeins}
\usepackage{comment}

\usepackage[linesnumbered,vlined,ruled,noend]{algorithm2e}
\usepackage{algorithm2e}

\begin{document}
\title{Efficient-Empathy: Towards Efficient and Effective Selection of Empathy Data}

\author{Linzhuang Sun$^{\dagger\S}$, Hao Liang$^{\dagger\ddagger}$, Jingxuan Wei$^{\S}$, Linkun Sun$^{\diamond}$, Bihui Yu$^{\S}$, Bin Cui$^{\ddagger*}$, Wentao Zhang$^{\ddagger*}$}
\affiliation{
$^\dagger$Equal Contribution~~~~~$^\ddagger$Peking University~~~~~$^\S$University of Chinese Academy of Sciences \newline $^\diamond$ Institute of Information Engineering, Chinese Academy of Sciences $^*$ Corresponding Author
 }
\affiliation{
$^\dagger$sunlinzhuang21@mails.ucas.ac.cn, $^\dagger$hao.liang@stu.pku.edu.cn, \{bin.cui, wentao.zhang\}@pku.edu.cn
}

\begin{abstract}
In recent years, with the rapid advancements in large language models (LLMs), achieving excellent empathetic response capability has become a crucial prerequisite. Consequently, managing and understanding empathetic datasets has gained increasing importance. However, empathetic data are typically trained without any quality selection, leading to inefficient data usage and wasted computational resources. Additionally, using raw data can result in low performance in empathetic dialogues. In this work, we present Efficient-Empathy, a sensibility and rationality score-based data selection algorithm that automatically selects sensibility and rationality data while discarding low-quality data. With only the sensibility data (59\% of the full dataset), our trained sensibility model efficiently achieves state-of-the-art (SoTA) performance. Furthermore, with multiple data selection hyperparameters, the sensibility model demonstrates SoTA performance, showcasing the robustness of our method. By integrating sensibility and rationality data with a MoE structure, we achieve even higher performance, demonstrating the effectiveness of our Efficient-Empathy algorithm.
\end{abstract}

\maketitle

\pagestyle{\vldbpagestyle}
\begingroup\small\noindent\raggedright\textbf{PVLDB Reference Format:}\\
\vldbauthors. \vldbtitle. PVLDB, \vldbvolume(\vldbissue): \vldbpages, \vldbyear.\\
\href{https://doi.org/\vldbdoi}{doi:\vldbdoi}
\endgroup
\begingroup
\renewcommand\thefootnote{}\footnote{\noindent
This work is licensed under the Creative Commons BY-NC-ND 4.0 International License. Visit \url{https://creativecommons.org/licenses/by-nc-nd/4.0/} to view a copy of this license. For any use beyond those covered by this license, obtain permission by emailing \href{mailto:info@vldb.org}{info@vldb.org}. Copyright is held by the owner/author(s). Publication rights licensed to the VLDB Endowment. \\
\raggedright Proceedings of the VLDB Endowment, Vol. \vldbvolume, No. \vldbissue\ %
ISSN 2150-8097. \\
\href{https://doi.org/\vldbdoi}{doi:\vldbdoi} \\
}\addtocounter{footnote}{-1}\endgroup

\ifdefempty{\vldbavailabilityurl}{}{
\vspace{.3cm}
\begingroup\small\noindent\raggedright\textbf{PVLDB Artifact Availability:}\\
The source code, data, and other artifacts will be released after the paper is accepted.
\endgroup
}

\section{Introduction}
\label{sec:intro}

In recent years, with the rapid advancements in large language models (LLMs)~\cite{chatgpt, llama}, data management has become a crucial aspect of these technologies~\cite{fernandez2023large, trummer2023bert, chen2023lingua, miao2024demystifying, nie2023flexmoe}. At the same time, \citet{bai2024survey} also demonstrates that data processing, selection, and management can significantly influence the performance of LLMs.

Empathy, the ability to understand and share the feelings of another, is a critical component of human social interaction and communication. It allows individuals to connect with others on an emotional level, fostering relationships and promoting prosocial behavior ~\cite{davis1983measuring}. In the future of Human-central AGI, excellent empathetic response capability is a crucial prerequisite, which focuses on equipping LLMs to understand and respond appropriately to human emotions ~\cite{rashkin2018towards}.

In previous research, empathy modeling has predominantly concentrated on shallow emotional labels ~\cite{wang2022empathetic, fu2023core, yang-etal-2023-exploiting-emotion, yufeng2024ctsm} and integrating information through modifications to the model architecture ~\cite{ghosal2020cosmic, zhou2021probing, sabour2022cem}. While the importance of model architectures is well acknowledged, the quality of data is also critical ~\cite{chen2023alpagasus, xu2023rethinking}. Additionally, \citet{fernandez2023large} indicates that new data management methods are needed for LLMs due to their massive data requirements. Previous empathetic studies underscore the significance of data-centric approaches, they often overlook efficient and robust data management, leading to three key challenges:

\begin{figure}
\centering 
\includegraphics[width=0.47\textwidth]{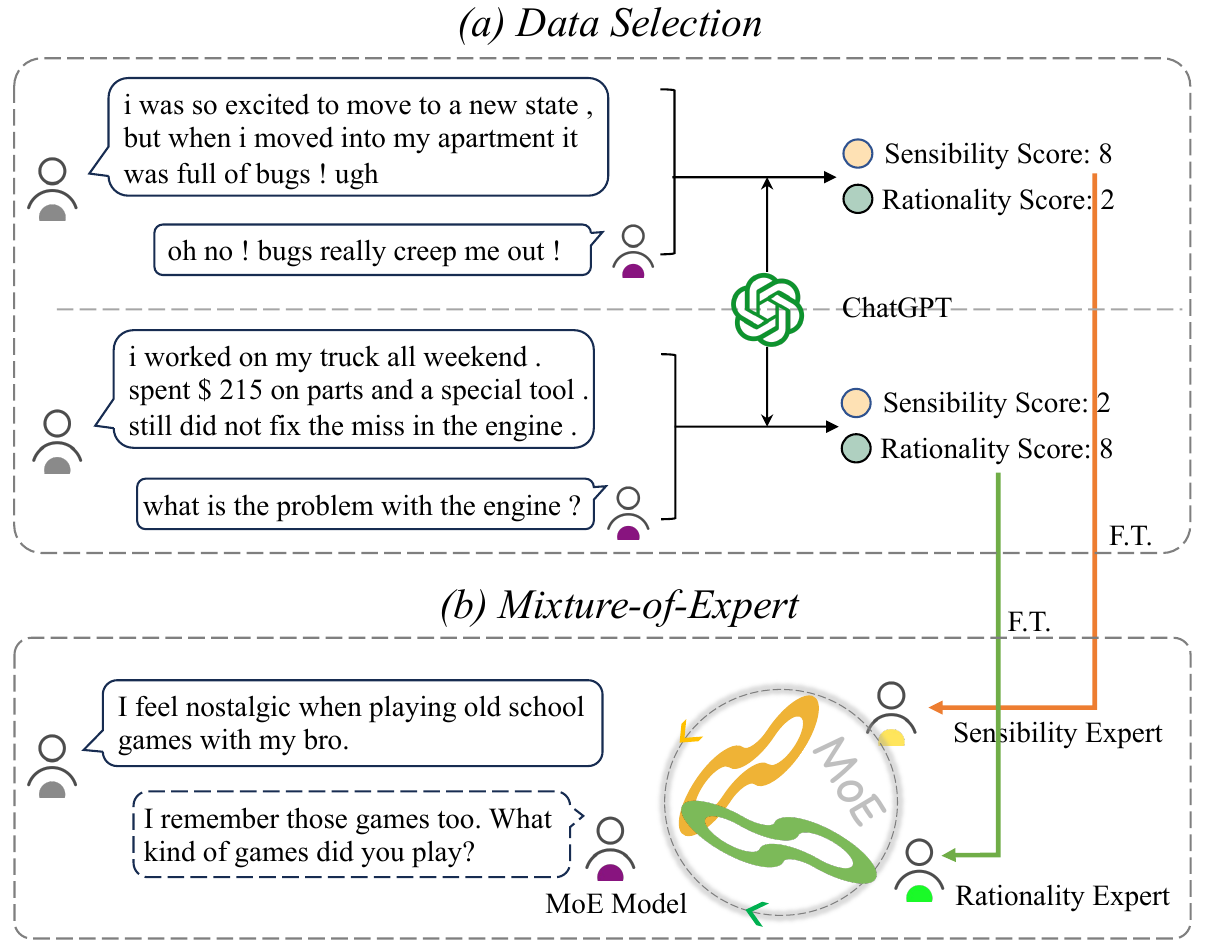} 
\caption{The pipeline of our approach (a) The data selection method utilized for classifying sensibility and rationality conversation. (b) Utilize sensibility and rationality data for MoE training}
\label{Fig.demo}
\end{figure}

\textbf{C1. Low Efficiency.}
Model-centric approaches\cite{sabour2022cem, yang-etal-2023-exploiting-emotion} neglect the distribution and quality of the data. They choose to fine-tune models using the entire dataset. However, some low-quality data may result in more training time and computation cost.

\textbf{C2. Low Robustness.} 
Prompt-based methods~\cite{qian2023harnessing} have enhanced the empathetic response capabilities of models by carefully designing prompts to elicit more contextual knowledge. However, these techniques are often specific to particular types of LLMs, limiting their generalizability and robustness.

\textbf{C3. Poor Effectiveness.}
Previous research~\cite{lin2019moel, ghosal2020cosmic, inproceedings, sabour2022cem, wang2022empathetic, zhou2022case} never examined the distribution of empathetic data, overlooking the roles of sensibility and rationality. Although Lamb~\cite{sun2023rational} indicates that sensibility and rationality positively impact empathy, it remains uncertain whether sensibility and rationality can be leveraged to select data and further enhance empathetic performance.

To address these issues, as shown in Figure \ref{Fig.demo}(a), we propose a new empathetic data selection method. We utilize LLMs to automatically rate the sensibility and rationality scores of empathetic data. Then, we select and discard data based on these scores. By using only the selected sensibility data, we can achieve SoTA performance by fine-tuning LLMs, which is highly data efficient. Additionally, we use multiple data selection thresholds to test the robustness of our method. The results in Table \ref{tab:threshold} show that our method consistently outperforms full data fine-tuning, demonstrating the robustness of our model. Furthermore, we utilize both the selected sensibility and rationality data to train a sensibility expert and a rationality expert. The improvements of As shown in \ref{Fig.demo}(b), we then further train a Mixture-of-Experts (MoE) model~\cite{jacobs1991adaptive}, achieving even higher SoTA performance, which demonstrates the effectiveness of our data selection method.
\begin{figure}
\centering 
\includegraphics[width=0.47\textwidth]{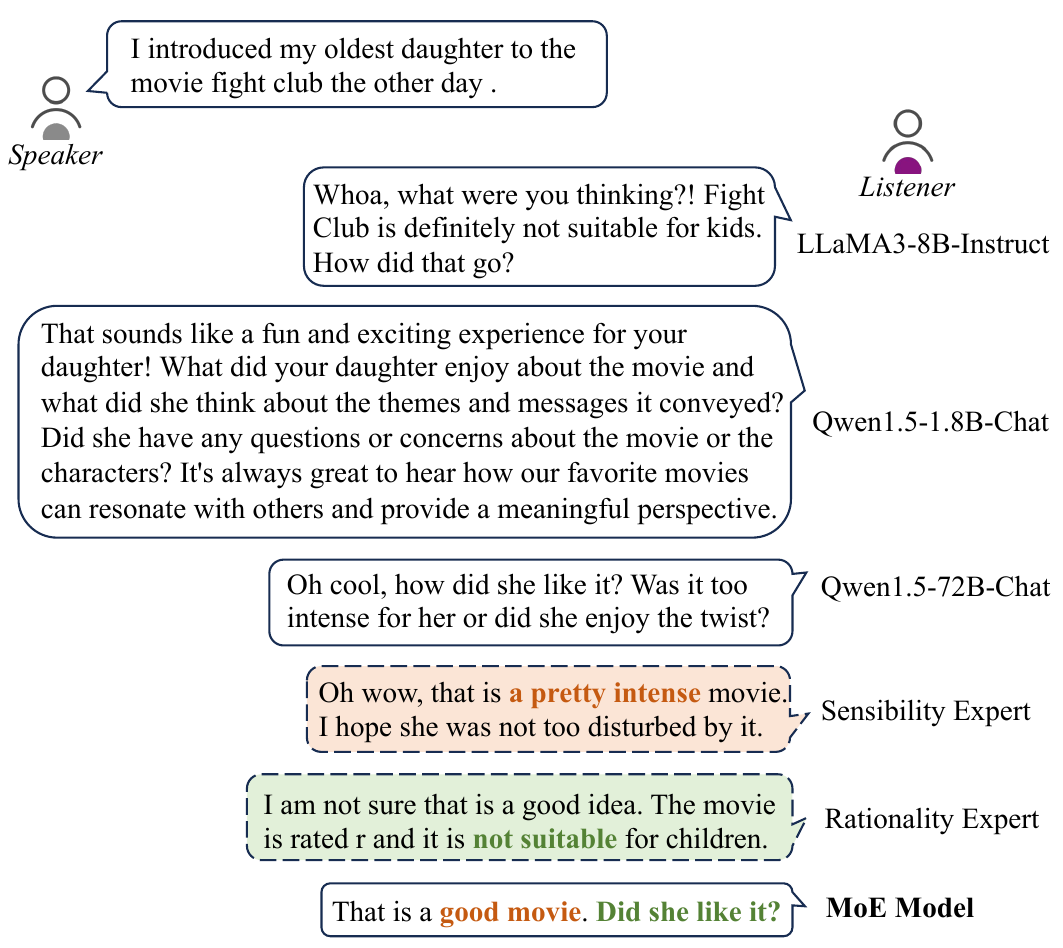} 
\caption{Comparison of Empathetic Responses from Different Models. Sensibility, Rationality, and MoE models are trained using data selected by Efficient-Empathy.}
\label{Fig.demo_response}
\end{figure}
The core contributions of this paper are summarized as follows:

\begin{itemize}
\item \textbf{New Perspective.}
Low data efficiency and low robustness are significant impediments to the practical adoption of empathy models. To the best of our knowledge, this study represents the first attempt to address these challenges from a data selection perspective.
\item \textbf{New Method.}
We propose a new data selection method for empathy, introducing the first sensibility and rationality-based data selection framework. Utilizing our meticulously curated sensibility and rationality scores, we pioneer the integration of sensibility and rationality data with a MoE model. Our method, which combines curated selected data with the MoE model, enables efficient, robust, and effective management of empathy data.
\item \textbf{SoTA Performance.} 
    \textbf{(1) \textit{High Efficiency.}} From a data-driven perspective, we analyze the distribution of sensibility and rationality cognition in empathetic dialogues. With a carefully curated dataset, our sensibility expert model outperforms the baseline using only 59\% of the data.
    \textbf{(2) \textit{High Robustness.}} As shown in Table \ref{tab:threshold}, with multiple data selection thresholds, our method consistently outperforms full data fine-tuning, demonstrating the robustness of our model.
    \textbf{(3) \textit{Effectiveness in Empathetic Response.}} Using only 59\% of the selected sensible data, our fine-tuned model achieved SoTA performance. Furthermore, we utilize both the selected sensibility and rationality data to train a MoE model, achieving even higher SoTA performance, which demonstrates the effectiveness of our data selection method.
\end{itemize}



\section{Related Work}

\subsection{Empathetic Response Generation}
 Imbuing models with empathy to enhance emotional experience is a crucial aspect of AGI. ~\citet{rashkin2018towards} constructed the ED dataset, the most authoritative dataset for measuring the empathic abilities of models, which has since become a benchmark for empathy research. Building on this dataset, numerous researchers have focused on improving models' empathy comprehension by integrating emotional labels and external knowledge.

~\citet{sabour2022cem} introduces commonsense knowledge inference into this task by enriching historical conversation data with the pre-trained COEMT model~\cite{bosselut2019comet}. Diverged from this approach, ~\citet{li2022knowledge} incorporate external knowledge through Graph Neural Networks (GNN) for context encoding. Additionally, \citet{wang2022empathetic} posit that using detailed sentiment labels can improve the accuracy of capturing user sentiment and puts forward a sentiment loss mechanism with multiple levels of granularity to enhance model training.  Moreover, ~\citet{kim2022emp} strives to provide empathetic responses by analyzing context at the word level. ~\citet{chen2022wish} employed a emotion detection algorithm based on psychological principles to pinpoint important statements in conversations. ~\citet{zhao2022don} take a different approach by not only recognizing the emotions of others but also assessing saved model emotional state.  Furthermore, ~\citet{qian2023think} divides the empathetic response task into two steps: verifying the semantic content and infusing emotional expression.  

In the LLM era, many researchs explore ways to boost the empathic potential of models through meticulous design of prompts~\cite{qian2023harnessing, wang2023enhancing, yang2024iterative}. Differently, ~\citet{sun2023rational} delves into the general sensibility and rationality, exploring their respective contributions to empathy. However, the importance of fine-grained sensibility and rationality cognition for empathy still lacks comprehensive investigation. 

\subsection{Data Quality and Data Selection}
The advent of large language models has brought about a substantial increase in the volume of training data.~\cite{llama, openai2023gpt} In this scenario, the quality and quantity of data become paramount. LLMs, trained on vast amounts of data, can capture subtle nuances and complex patterns in language, excelling in various natural language processing tasks. However, the increase in data volume also brings new challenges, particularly in data management, cleaning, and annotation.~\cite{bai2024survey} In this section, we mainly discuss the effectiveness of data quality and data selection.

\paragraph{Data Quality}: High-quality data can significantly enhance the performance of models.~\cite{llama3repo} As the volume of data increases, ensuring high data quality becomes more challenging because it requires more resources for data cleaning, selection and annotation.~\cite{bai2024survey} Poor quality data can lead to models learning incorrect patterns and making inaccurate predictions. 

\paragraph{Data Selection}: 
LLMs-based methods were commonly used in data selection.~\cite{bai2024survey} For instance, \citet{du2023mods} leverages DeBERTa~\cite{he2020deberta} for scoring, retaining high-quality data, and combining it with the k-center greedy algorithm to select diverse data. \citet{chen2023alpagasus} score the accuracy of data using ChatGPT to pick out high-quality data. \citet{xu2023rethinking} use GPT-4 to rewrite data to increase their complexity and then streamline it by reducing its variety and improving its quality. \citet{liu2023makes} train two models using ChatGPT's labeled data to score the quality and complexity of the data. \citet{lu2023instag} rely on ChatGPT to tag each instance, defining its complexity and diversity based on these tags. \citet{parkar2024selectllm} first cluster the data, and then use GPT-4 to select high-quality data for each cluster.

\subsection{Mixture of Expert}
The MoE framework is initially introduced by ~\citet{jacobs1991adaptive} with the aim of merging the functionalities of domain models. ~\citet{shazeer2017outrageously} suggestes the use of token-level routing and sparsely-gated mechanisms to enhance the robustness of expert model. Recently, as ~\citet{lepikhin2020gshard} and ~\citet{fedus2022switch} extend MoE to the transformer structure, a growing number of sophisticated routing mechanisms ~\cite{pan2024dense}, load balancing techniques~\cite{zhong2024lory}, and distributed training strategies ~\cite{gale2023megablocks} have been proposed.

Recently, the integration of MoE with LLMs has yielded significant advancements in general domains. Mixtral-8x22B ~\cite{jiang2024mixtral} combines eight 22B parameter models, achieving performance comparable to GPT. DeepSeekMOE ~\cite{dai2024deepseekmoe} enrich the expert configuration by implementing shared and fine-grained expert mechanisms. Qwen1.5-MoE-A2.7B ~\citet{qwen1.5}, utilizing the MoE approach, match the performance of a 7B parameter model while activating only one-third of its parameters. However, the application of MoE in the domain of empathetic responses remains largely unexplored.


\section{Efficient-Empathy Data Selection}

\begin{figure*}
\centering 
\includegraphics[width=1.0\textwidth]{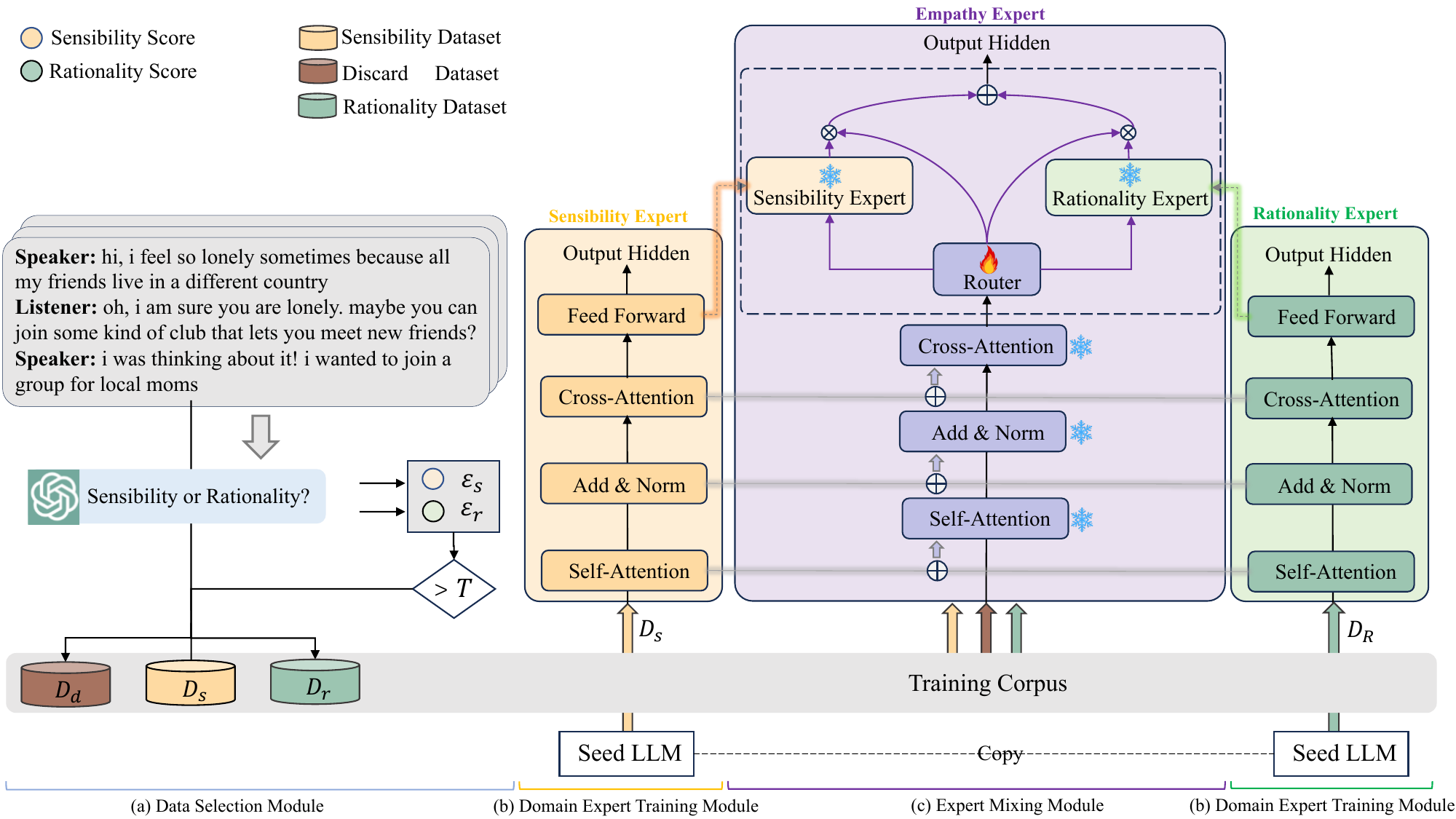} 
\caption{The overall pipeline of Efficient-Empathy consists of three parts: (a) the Data Selection Module, which classifies the empathetic dataset into sensibility, rationality, and discard datasets; (b) the Domain Expert Training Module, which uses the selected datasets to fine-tune LLMs and acquire sensibility and rationality experts; and (c) the Expert Mixing Module, which integrates the sensibility and rationality experts into the MoE empathy model.}
\vspace{2mm}
\label{Fig.model}
\end{figure*}

\subsection{ED Datasets}

Our study is based on the authoritative EmpatheticDialogues (ED) dataset~\cite{rashkin2018towards}, which consists of 25,000 daily conversations encompassing 32 uniformly distributed emotional labels. This dataset is meticulously curated through crowdsourcing, involving 810 workers from Amazon Mechanical Turk~\footnote{\url{https://www.mturk.com}}. Each conversation is constructed in a one-on-one format, pairing two workers. One worker assumes the role of the speaker, responding according to a given emotional label and situation, while the other worker assumes the role of the listener, providing empathetic responses to the speaker. On average, each conversation consists of 4.31 exchanges, with each exchange containing approximately 15.2 words.

\subsection{ED Data Selection}
We use the prompt shown in Figure \ref{Fig.prompt} to automatically evaluate the dataset. Specifically, we utilize ChatGPT to assess the sensibility and rationality of the ED dataset. Each dialogue in the dataset is assigned a sensibility score \( S \) and a rationality score \( R \).

Before evaluating the processed dataset, we perform data cleaning to ensure the accuracy of the GPT-evaluated scores. This includes correcting common typographical errors, such as fixing "Rationalality" to "Rationality".

After the data cleaning process, we start the selection process by establishing a threshold value \( T \) for both sensibility and rationality. Dialogues with rationality scores \( R \) less than the threshold \( T \) and sensibility scores \( S \) greater than the threshold \( T \) are included in the sensibility dataset \( D_s \), while dialogues with rationality scores \( R \) greater than the threshold \( T \) and sensibility scores \( S \) less than the threshold \( T \) are included in the discard dataset \( D_d \). All other dialogues are included in the rationality dataset \( D_r \). The process is illustrated on the left side of Figure \ref{Fig.model}.

Mathematically, the selection criteria can be defined as:

\begin{equation}
D_s = \{ d \in D \mid R(d) < T \ \text{and} \ S(d) > T \}
\end{equation}

\begin{equation}
D_d = \{ d \in D \mid R(d) > T \ \text{and} \ S(d) < T \}
\end{equation}

\begin{equation}
D_r = \{ d \in D \mid \neg (d \in D_s \cup d \in D_d) \}
\end{equation}

where \( D \) represents the original dataset, \( S(d) \) denotes the sensibility score of dialogue \( d \), and \( R(d) \) denotes the rationality score of dialogue \( d \).

In summary, our data selection process involves evaluating each dialogue for sensibility and rationality using ChatGPT, assigning scores accordingly, and then filtering the dialogues based on the established threshold \( T \) to create the sensibility dataset \( D_s \), the discard dataset \( D_d \), and the rationality dataset \( D_r \).

To summarize the contents of this section, we translated the data selection process into an algorithm, as shown in Algorithm \ref{alg:EDDataSelection}.

\begin{algorithm}
\SetAlgoLined
  \KwIn{Original ED dataset $D$, threshold $T$}
  \KwOut{Sensibility dataset $D_s$, Discard dataset $D_d$, Rationality dataset $D_r$}
  $D_s \gets \emptyset$ \;
  $D_d \gets \emptyset$ \;
  $D_r \gets \emptyset$ \;

  \For{each dialogue $d \in D$}{
    Assign sensibility score $S(d)$ using ChatGPT with prompt in Figure \ref{Fig.prompt}\;
    Assign rationality score $R(d)$ using ChatGPT with prompt in Figure \ref{Fig.prompt}\;
    \If{$R(d) < T$ and $S(d) > T$}{
      $D_s \gets D_s \cup \{d\}$ \;
    }
    \ElseIf{$R(d) > T$ and $S(d) < T$}{
      $D_d \gets D_d \cup \{d\}$ \;
    }
    \Else{
      $D_r \gets D_r \cup \{d\}$ \;
    }
  }
    \textbf{return} $D_s, D_d, D_r$
    \caption{ED Data Selection Process}
    \label{alg:EDDataSelection}
    \vspace{2mm}
\end{algorithm}

\subsection{Distribution of Rationality and Sensibility}
In this section, we provide an analysis of the distribution of rationality and sensibility scores. Each dialogue in the dataset is evaluated on a scale from 0 to 10 for both rationality and sensibility. To visualize the distribution of these scores, we plot a 2D histogram showing the frequency of each score from 0 to 10.

\begin{figure}[h]
    \centering
    \includegraphics[width=0.47\textwidth]{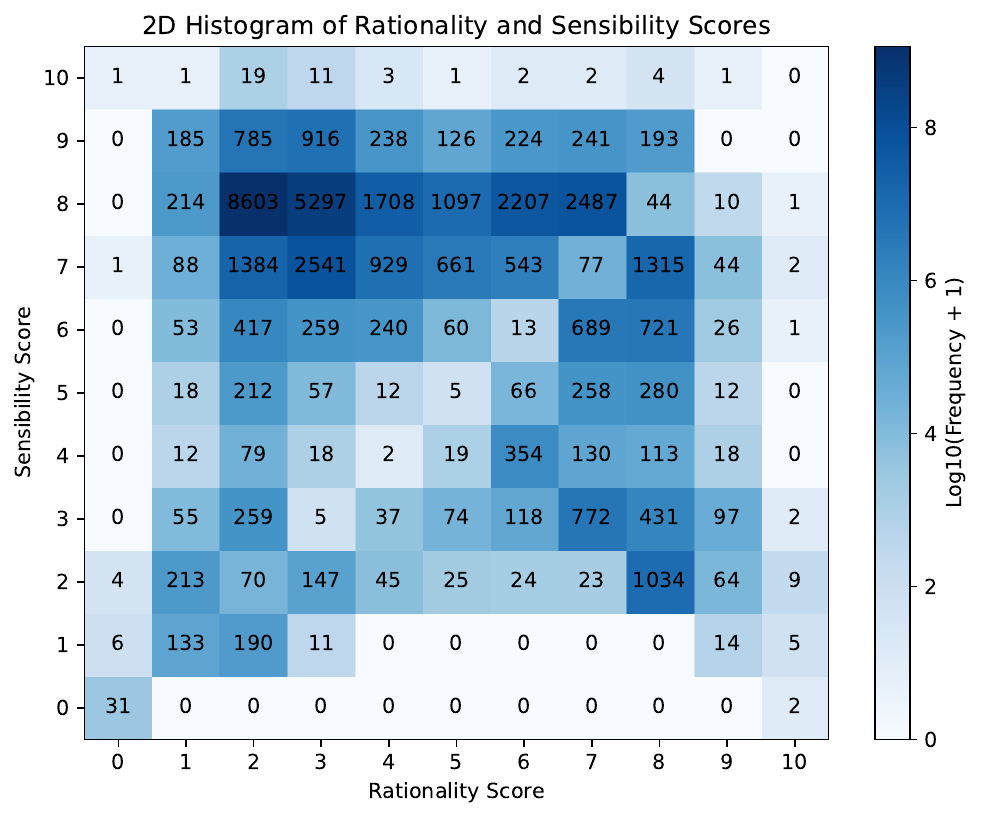}
    \caption{2D Histogram of Rationality and Sensibility Scores. The x-axis represents rationality scores, the y-axis represents sensibility scores, and the color intensity indicates the frequency of each combination of scores.}
    \label{Fig.distribution}
\end{figure}

Figure \ref{Fig.distribution} illustrates the 2D histogram for both rationality and sensibility scores. The x-axis represents the rationality scores, while the y-axis represents the sensibility scores. The color intensity indicates the frequency of each combination of scores, with a color bar on the right side providing the exact frequency values.

From the 2D histogram, we can observe several notable trends and patterns:
\begin{itemize}
\item \textbf{High-Frequency Clusters}: The highest frequency cluster is found at the combination of a rationality score of 2 and a sensibility score of 8, with a frequency of 8603. This indicates that dialogues with low rationality and high sensibility are common in the dataset. 
\item \textbf{General Trends}: There is a tendency for dialogues to score higher on sensibility compared to rationality. This is evidenced by the higher frequencies of scores in the upper part of the y-axis (sensibility scores) compared to the x-axis (rationality scores). This matches the ED dataset, which typically contains dialogues with high sensibility content.
\item \textbf{Balanced Scores}: Dialogues with balanced rationality and sensibility scores (e.g., both scores around 5) are less frequent. This highlights the rarity of dialogues that equally exhibit both logical reasoning and emotional depth, suggesting a potential area for improving dialogue generation models to better balance these attributes.
\item \textbf{Outliers}: Some outliers exist where dialogues have extremely high sensibility but very low rationality, and vice versa. These outliers are crucial for understanding edge cases and improving model robustness in handling diverse dialogue scenarios.
\end{itemize}

This visualization helps in understanding the overall distribution and identifying any trends or patterns in the data, which can be valuable for further analysis and improving the dialogue system.

\section{Empathetic MoE}
\label{section4}
In accordance with $D_s$ and $D_r$, we conduct single-expert training for sensibility and rationality abilities. Subsequently, expert models Efficient-Empathy are integrated using the Branch-Train-Mix (BTX) method~\cite{sukhbaatar2024branch}. The overall model structure is shown in Fig~\ref{Fig.model}. And the algorithm description is shown in Algorithm 2.

\subsection{Empathetic Domain Experts}
In this subsection, we transfer knowledge from LLM to create specialized expert models via Supervised Fine-Tuning (SFT). Specifically, we utilize LLaMA3-8B-Instruct as the seed model to engage in LoRA fine-tuning on two domain datasets, $D_s$ and $D_r$, thereby acquiring the sensibility expert $M_s$ and the rationality expert $M_r$ separately: 

\begin{equation}
\begin{split}
    M_s = \theta_{s}^{SFT}(LLM; D_s)
\end{split}
\end{equation}

\begin{equation}
\begin{split}
    M_r = \theta_{r}^{SFT}(LLM; D_r)
\end{split}
\end{equation}
The derived models, $M_s$ and $M_r$, undergo training to excel within their respective domains, establishing the foundation for the subsequent mixing stage.

\subsection{Empathetic MoE Architecture}
\label{sec:eme}
Building upon the insights of domain experts, we employ the MoE approach to incorporate them into a comprehensive empathetic model, denoted as $M_e$. Diverging from the Branch-Train-Merge (BTM) ~\cite{li2022branch} methodology, which exclusively consolidates the final feature representations, our approach introduces a collaborative configuration of Feed-Forward Network (FFN) layers. It further allocates decision-making weights to individual experts via a sophisticated soft routing system, thereby enhancing the model's selective empathy capabilities.

In a single expert model, each transformer block comprises an attention module and an FFN module. The structure of $M_e$ is similar, with the exception that the FFN layer is substituted by an MoE layer, which is a combination of multiple single expert FFNs. Particularly, in the $i$-th transformer block, the FFN layers of $M_s$ and $M_r$ are denoted as $FF_i^s(x)$ and $FF_i^r(x)$, respectively. The $i$-th MoE layer $MoE_i(x)$ is defined as follows:

\begin{align}
    MoE_i(x) &= Router_i(x) \cdot FF_i^s(x) \nonumber \\
             &\quad + (1 - Router_i(x)) \cdot FF_i^r(x),
\end{align}
where 
\begin{equation}
\begin{split}
    Router_i(x) = SoftMax(W_i x + b_i), 
\end{split}
\end{equation}
$Router_i$ serves as a soft routing mechanism that output values ranging between 0 and 1 to regulate the influence of domain experts.

$W_i$ and $b_i$ is the linear transformation and bias of the $i$-th MoE layer. For the parameters and weights of other layers in the Efficient-Empathy model, we initialize them using the average weights of the corresponding layers in the expert models:

\begin{equation}
M_e^i(x) =
\begin{cases} 
\text{MoE}_i(x) & \text{MoE Layer} \\
\text{Average}(M_s^i(x), M_r^i(x)) & \text{Other Layer}
\end{cases}
\end{equation}

After constructing the model, we introduce new random initialized router parameters and then average the weights of domain experts. Consequently, a second stage of training is conducted in the subsequent experiment to activate the overall parameter weights of the empathy model.

\begin{algorithm}[t]
  \caption{Empathy MoE Training Process}\label{alg:MoETraining}
  \KwIn{Sensibility dataset $D_s$, Rationality dataset $D_r$, Discard dataset $D_d$, Seed model $\text{LLM}$, Sensibility expert $M_s$, Rationality expert $M_r$, Routing mechanism $\text{Router}$, Feed-Forward Network $\text{FFN}$}
  \KwOut{Empathy Model $M_e$}
  
  $M_s \gets \text{SFT}(\text{LLM}; D_s)$\;
  $M_r \gets \text{SFT}(\text{LLM}; D_r)$\;
  $M_e \gets \emptyset$ \;

  \For{$layer_i \in M_s$}{
    \If{$layer_i \in \text{FFN}$}{
      $M_e^{i} = \text{Router} \cdot M_s^{i}+ (1- \text{Router}) M_r^{i}$\;
    }
    \Else{
      $M_e^{i} = \text{Average}(M_s^{i}, M_r^{i})$\;
    }
  }
  $M_e \gets \text{SFT}(M_e; D_r, D_s, D_n)$\;
\Return $M_e$
\end{algorithm}

\subsection{Empathetic MoE Training}


In the empathetic response generation task, given a conversation message $C=[S_1, L_1, S_2, ..., S_N]$ of length $N$, where $S$ is Speaker's utterance and $L$ is Listener's utterance. Our model then generate a response $R=[r_1, r_2, ..., r_m]$ based on emotional expert and rational expert in context with $C$, where $m$ is the length of the token sequence:
\begin{figure*}
\centering 
\includegraphics[width=1.0\textwidth]{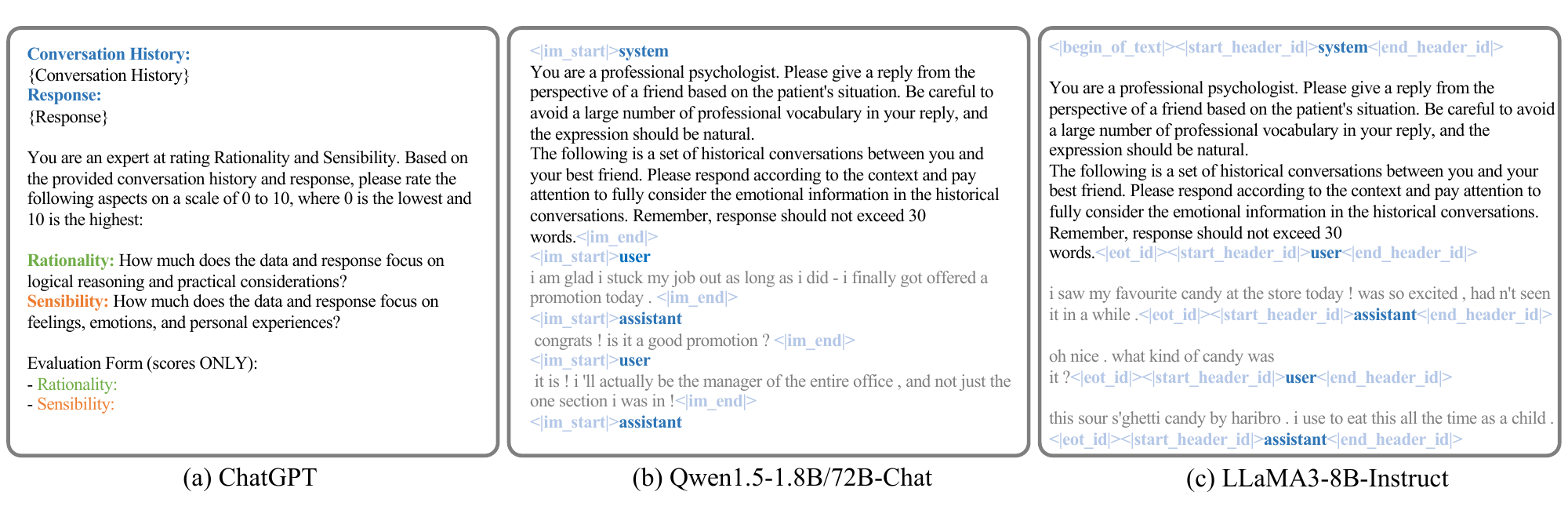} 
\caption{Meticulously designed prompts for Data Evaluation and empathetic response generation.}
\label{Fig.prompt}
\end{figure*}

\begin{equation}
\begin{split}
    R = MoE(C)
\end{split}
\end{equation}

The training loss is the standard negative loglikelihood (NLL) loss on the generated response $R$:

\begin{equation}
    \mathcal{L}_{nll} = - \sum_{t=1}^{m}\log(r|C, r_{<t}) 
\end{equation}

After completing the domain expert training and initializing the MoE model, we will provide a detailed description of the model training process in Section~\ref{sec:moe-train}.

\section{Experiments}
In this section, we first introduce the experimental setups, including data selection and the training process. We then aim to answer the following questions to verify the effectiveness, efficiency, and robustness of our proposed Efficient-Empathy model:\\
\textbf{Q1}: Can our model achieve SoTA performance with the sensibility and rationality dataset selected by Efficient-Empathy? \\
\textbf{Q2}: Can we achieve SoTA performance efficiently by only utilizing the sensibility data? \\
\textbf{Q3}: Can our model robustly and consistently achieve SoTA performance with varying data selection thresholds? \\
\textbf{Q4}: Can our model achieve good human evaluation results, given the importance of human evaluation for empathy? \\
\textbf{Q5}: Can we visualize our results with well-analyzed case studies?
\textbf{Q6}: Is the SoTA performance achieved due to our MoE structure or the Efficient-Empathy data selection method? \\

\subsection{Experimental Setup}

\subsubsection{Datasets}
We utilize the ED dataset with three Efficient-Empathy selected subsets: Sensibility, Rationality, and Neutral.

\subsubsection{Models}
We employ four commonly used LLMs: Qwen1.5-1.8B-Chat, LLaMA2-13B-Instruct, LLaMA3-8B-Instruct, Qwen1.5-72B-Chat. The models are trained using HuggingFace's Trainer framework. During the experiments, we use the DeepSpeed library with AdamW as the optimizer and set zero\_optimization to 3. For LoRA-Tuning, we use a rank of 8, an alpha value of 32, a dropout rate of 0.1, and target specific modules, including q\_proj, k\_proj, v\_proj, o\_proj, gate\_proj, up\_proj, and down\_proj.

\subsubsection{MoE Training}
\label{sec:moe-train}
To develop a model with comprehensive empathetic capacity, we employ domain-specific experts using the MoE approach. Initially, we create two duplicates of the seed model to train sensibility and rationality experts in parallel. After completing the training of these experts, we integrate them into a MoE model using the method described in Section~\ref{section4}. Subsequently, all parameters of the empathy model, except the router, are kept frozen while we perform further fine-tuning on the entire training dataset. This training phase aims to enable the router to distinguish between tokens with disparate cognitive attributes and assign corresponding expert weights, while preventing the overfitting of sub-experts.

\subsubsection{Baselines}
We compare our model with the following baselines:

1. \textbf{MoEL}~\cite{lin2019moel}: Creates a decoder for each emotion to generate a final response.
2. \textbf{MIME}~\cite{ghosal2020cosmic}: Simulates user emotions and generates empathetic responses by introducing randomness.
3. \textbf{EmpDG}~\cite{inproceedings}: Includes an empathetic information generator and a sentiment discriminator.
4. \textbf{CEM}~\cite{sabour2022cem}: Incorporates the COMET pre-trained model for common sense knowledge in empathetic response generation.
5. \textbf{SEEK}~\cite{wang2022empathetic}: Focuses on sentence-level sentiment information using attention mechanisms.
6. \textbf{CASE}~\cite{zhou2022case}: Utilizes external resources COMET and ConceptNet to enhance cognitive and emotional abilities.
7. \textbf{E-CORE}~\cite{fu2023core}: Explores intrinsic sentiment through emotion correlation learning and supervision.
8. \textbf{KEMP}~\cite{li2022knowledge}: Uses ConceptNet and VRC-NED as external knowledge sources for contextual modeling.
9. \textbf{CAB}~\cite{gao2023cab}: Divides empathy response generation into cognition, affection, and behavior.
10. \textbf{ESCM}~\cite{yang-etal-2023-exploiting-emotion}: Uses dynamic emotion-semantic vectors and dependency trees to guide empathetic response generation.
11. \textbf{DCKS}~\cite{cai2023improving}: Incorporates an adaptive module for commonsense knowledge selection to ensure consistency.
12. \textbf{CTSM}~\cite{yufeng2024ctsm}: Categorizes emotions into fine-grained trait and state emotions to improve sentiment perception.
13. \textbf{Lamb}~\cite{sun2023rational}: Enhances the empathetic response capability by jointly utilizing self-presentation theory and chain-of-thought data from LLaMA2-13B-Chat.

\subsubsection{Evaluation}
Following previous works~\cite{lin2019moel, ghosal2020cosmic, inproceedings, sabour2022cem, wang2022empathetic, zhou2022case, fu2023core, li2022knowledge, gao2023cab, yang-etal-2023-exploiting-emotion, cai2023improving, sun2023rational}, we evaluate the performance of our model using both automatic and human evaluation metrics to provide a comprehensive assessment of its capabilities. As illustrated in Figure~\ref{Fig.prompt}, we use a meticulously designed prompt for LLM inference.

\textbf{Automatic Evaluation Metrics}: We use corpus-level BLEU
(B-1 to B-4), sentence-level ROUGE (R-1, R-2), and Distinct (Dist-1, Dist-2) as automatic evaluation metrics. BLEU and ROUGE scores quantify the resemblance between the generated text and the ground-truth text, with higher scores indicating greater likeness. Distinct-N evaluates the diversity of the content, with higher values suggesting a wider range of diverse representations. The perplexity metric is not utilized as it measures confidence in the generated sentences, which is not specific to empathy scenarios. Given the absence of authoritative literature demonstrating that LLMs surpass humans in empathy judgment, we rely on scientific human evaluation to assess the methods' effectiveness.

\textbf{Human Evaluation Metrics}: We select Coherence, Empathy, Information, and Continuity as human evaluation metrics.
\begin{itemize}
    \item \textbf{Coherence}: Evaluates the correspondence between the text produced by the model and the desired response.
    \item \textbf{Empathy}: Assesses the model's ability to understand the speaker's situation and effectively express concern.
    \item \textbf{Informativeness}: Gauges the amount of information present in the generated responses.
    \item \textbf{Continuity}: Reflects the model's capability to sustain the conversation.
\end{itemize}
We conduct an A/B test to compare the effectiveness of our model against several baselines. Specifically, we randomly select 200 examples from the test dataset. For each instance, the context is paired with two responses: one generated by our model and the other by a baseline model. Three evaluators assess each pair of responses and determine a winner, a loser, or a tie based on the four dimensions: Coherence, Empathy, Information, and Continuity.
\begin{table*}[htbp]
  \centering
  \caption{Results of the automatic evaluation on baseline models, the sensibility model, and the MoE model are presented. The best performance is highlighted in bold, and the purple table represents the increased values. 
  }
  \resizebox{0.8\linewidth}{!}{
    \begin{tabular}{ccccccccc}
    \toprule
    \textbf{Models} & \textbf{B-1} & \textbf{B-2} & \textbf{B-3} & \textbf{B-4} & \textbf{R-1} & \textbf{R-2} & \textbf{Dist-1} & \textbf{Dist-2} \\
    \midrule
    MoEL~\cite{lin2019moel}  & 18.07 & 8.30  & 4.37  & 2.65  & 18.24 & 4.81  & 0.59  & 2.64 \\
    MIME~\cite{ghosal2020cosmic}  & 18.60 & 8.39  & 4.54  & 2.81  & 17.08 & 4.05  & 0.47  & 1.66 \\
    EmpDG~\cite{inproceedings} & 19.96 & 9.11  & 4.74  & 2.80  & 18.02 & 4.43  & 0.46  & 1.99 \\
    CEM~\cite{sabour2022cem}   & 16.12 & 7.29  & 4.06  & 2.03  & 15.77 & 4.50  & 0.62  & 2.39 \\
    SEEK~\cite{wang2022empathetic}  & 10.77 & 4.40  & 2.02  & 1.08  & 12.74 & 2.94  & 0.68  & 2.81 \\
    CASE~\cite{zhou2022case}  & 15.59 & 7.22  & 3.80  & 2.24  & 17.33 & 4.67  & 0.65  & 3.37 \\
    E-CORE~\cite{fu2023core} & -   & -   & -   & -   & -   & -   & 0.72  & 3.49 \\
    KEMP~\cite{li2022knowledge}  & 16.72 & 7.17  & 3.77  & 2.33  & 16.11 & 3.31  & 0.66  & 3.07 \\
    CAB~\cite{gao2023cab}   & 19.23 & 8.55  & 4.36  & 2.57  & 17.50 & 4.13  & 1.13  & 4.23 \\
    ESCM~\cite{yang-etal-2023-exploiting-emotion}  & -   & -   & -   & -   & -   & -   & 1.19  & 4.11 \\
    DCKS~\cite{cai2023improving}  & 18.75 & 9.12  & 5.38  & 3.57  & 19.14 & 5.45  & 1.57  & 6.02 \\
    CTSM~\cite{yufeng2024ctsm}  & -   & -   & -   & -   & -   & -   & 2.00  & 7.34 \\
    Lamb~\cite{sun2023rational}  & \textbf{22.00} & \textbf{10.49} & \textbf{6.07}  & \textbf{3.97}  & \textbf{19.55} & \textbf{5.47}  & \textbf{1.80}  & \textbf{7.73} \\
    \midrule
    Qwen1.5-1.8B-Chat~\cite{qwen1.5}  & 10.43  & 3.50  & 1.59  & 0.84  & 12.95 & 1.64  & 2.41  & 17.98 \\
    LLaMA2-13B-Instruct~\cite{touvron2023llama}   & 11.69 & 4.03  & 1.79  & 0.93  & 13.27 & 1.83  & 2.91  & 18.92 \\
    LLaMA3-8B-Instruct~\cite{touvron2023llama}   & 13.17 & 4.42  & 1.92  & 1.02  & \textbf{14.12} & 1.68  & 2.69  & 18.70 \\
    Qwen1.5-72B-Chat~\cite{qwen1.5}   & \textbf{14.19} & \textbf{4.85}  & \textbf{2.27}  & \textbf{1.23}  & 13.83 & \textbf{1.97}  & 3.29  & 22.68 \\

    \midrule
    \textbf{Sensibility} & \boxed{22.34} & \boxed{11.25} & \boxed{6.58}  & \boxed{4.21}  & \boxed{19.82} & \boxed{5.79}  & \boxed{3.00}     & \boxed{15.44}  \\
    \rowcolor{Violet!15} 
    \textit{+ Compared with Lamb} & $\blacktriangle$ 0.34   & $\blacktriangle$ 0.76  & $\blacktriangle$ 0.51  & $\blacktriangle$ 0.24  & $\blacktriangle$ 0.27  & $\blacktriangle$ 0.32  & $\blacktriangle$ 1.2  & $\blacktriangle$ 7.71 \\
    \rowcolor{Violet!15} 
    \textit{+ Compared with Qwen1.5-72b} & $\blacktriangle$ 8.15   & $\blacktriangle$ 6.40  & $\blacktriangle$ 4.31  & $\blacktriangle$ 2.98  & $\blacktriangle$ 5.70  & $\blacktriangle$ 3.82  & -  & - \\
    
    \textbf{MoE Model} & \boxed{\textbf{23.04}} & \boxed{\textbf{11.62}} & \boxed{\textbf{6.68}} & \boxed{\textbf{4.22}} & \boxed{\textbf{20.28}} & \boxed{\textbf{6.15}} & \boxed{2.34}  & \boxed{10.91} \\
    \rowcolor{Violet!15} 
    \textit{+ Compared with Lamb} & $\blacktriangle$ 1.04   & $\blacktriangle$ 1.13  & $\blacktriangle$ 0.61  & $\blacktriangle$ 0.25  & $\blacktriangle$ 0.73  & $\blacktriangle$ 0.68  & $\blacktriangle$ 0.54  & $\blacktriangle$ 3.18 \\
    \rowcolor{Violet!15} 
    \textit{+ Compared with Qwen1.5-72b} & $\blacktriangle$ 8.85   & $\blacktriangle$ 6.77  & $\blacktriangle$ 4.41  & $\blacktriangle$ 2.99  & $\blacktriangle$ 6.16  & $\blacktriangle$ 4.18  &  -  & -  \\
    \bottomrule
    \end{tabular}%
    }
  \label{tab:main_exp}%
\end{table*}%

\subsubsection{Settings.}
For the four LLMs used in our experiment, we use the hyperparameters from the official repositories: LLaMA3-8B-Instruct\footnote{\url{https://huggingface.co/meta-llama/Meta-Llama-3-8B-Instruct}}, Qwen1.5-1.8B-Chat\footnote{\url{https://huggingface.co/Qwen/Qwen1.5-1.8B-Chat}}, LLaMA2-13B-Chat\footnote{\url{https://huggingface.co/meta-llama/Llama-2-13b-chat-hf}}, and Qwen1.5-72B-Chat\footnote{\url{https://huggingface.co/Qwen/Qwen1.5-72B-Chat}}. All experiments are conducted on an 8*A100 NVIDIA GPU machine with a 120-core CPU and 960GB of memory.

\subsection{Main Experiments}
\label{sec:main_exp}

To answer \textbf{Q1}, we compare our Sensibility and MoE model against the baselines in Table~\ref{tab:main_exp}.

\textbf{Sensibility:} Following the method outlined in Section, we performed SFT training using the $D_s$ dataset based on the LLaMA3-8B-Instruct model. 

\textbf{MoE Model:} The domain experts consist of Sensibility and Rationality. The \textbf{Rationality} expert, similar to the Sensibility expert, is obtained through SFT training using the $D_r$ dataset based on the LLaMA3-8B-Instruct model. Subsequently, following the approach described in Section~\ref{sec:moe-train}, further training is conducted on the entire dataset ED.

Our model demonstrates significant improvements over the baselines in BLEU
(B-1 to B-4), ROUGE (R-1, R-2), and Distinct (Dist-1, Dist-2) scores, highlighting its effectiveness. Similarly, when compared with the LLM, our approach achieves substantial enhancements in BLEU
(B-1 to B-4), ROUGE (R-1, R-2) metrics. 
Interestingly, although LLMs exhibit high Distinct (Dist-1, Dist-2) scores, the lower performance in BLEU (B-1 to B-4), ROUGE (R-1, R-2) suggests a lack of quality in their generated responses.

The discrepancy between BLEU and Distinct metrics can be attributed to their fundamental differences. Distinct-N measures the diversity of the generated text by calculating the ratio of unique n-grams, promoting lexical variety and creativity. In contrast, BLEU-n and ROUGE-n assess the n-gram overlap between the generated text and reference texts, focusing on precision and recall, respectively. High Distinct-N scores indicate a broad range of expressions, which, while enhancing diversity, reduce the likelihood of exact n-gram matches with the reference texts, thereby resulting in lower BLEU-n and ROUGE-n scores. This highlights the inherent trade-off between achieving diversity and maintaining overlap with reference texts in text generation tasks. Given that the Distinct metric of our model significantly surpasses that of the baselines, we have opted not to further analyze the distinctiveness of LLMs.

\begin{table*}[htbp]
  \centering
  \caption{The Full F.T. model utilizes 40,250 training data instances, whereas the Sensibility model employs only 23,862 (59\%) training data instances. Sensibility data can efficiently and effectively improve empathetic performance.}
  \resizebox{0.8\linewidth}{!}{
    \begin{tabular}{cccccccccccccc}
    \toprule
          & \textbf{\#Data} & \textbf{Data Percentage} & \textbf{Models} & \textbf{B-1} & \textbf{B-2} & \textbf{B-3} & \textbf{B-4} & \textbf{R-1} & \textbf{R-2}  \\
    \midrule
    \multirow{2}[1]{*}{LLaMA3-8B-Instruct} 
          & 40,250 & 100\% & Full F.T. & 21.23 & 10.40 & 5.98  & 3.84  & 19.49  & 5.58  \\
          & 23,862 & 59\%  & \textbf{Sensibility} & \textbf{22.34} & \textbf{11.25} & \textbf{6.58}  & \textbf{4.21}  & \textbf{19.82} & \textbf{5.79}\\
    \midrule
    \multirow{2}[1]{*}{Qwen1.5-7B-Chat} 
          & 40,250 & 100\% & Full F.T. & 20.84 & 9.19  & 4.89  & 2.90   & 17.44 & 4.09 \\
          & 23,862 & 59\%  & \textbf{Sensibility} & \textbf{21.34} & \textbf{10.27} & \textbf{5.82}  & \textbf{3.66}  & \textbf{18.86} & \textbf{5.28}  \\
    \midrule
    \multirow{2}[2]{*}{Qwen1.5-1.8B-Chat} 
          & 40,250 & 100\% & Full F.T. & 17.45 & 6.78  & 3.22  & 1.76  & 14.89 & 3.08  \\
          & 23,862 & 59\%  & \textbf{Sensibility} & \textbf{18.2}  & \textbf{7.50}   & \textbf{3.73}  & \textbf{2.16}  & \textbf{15.74} & \textbf{3.48}  \\
    \bottomrule
    \end{tabular}%
    }
  \label{tab:sensibility_effect}%
\end{table*}%


\begin{table*}[htbp]
  \centering
  \caption{Performance of Our model Across Different Data Selection Thresholds}
  \resizebox{0.8\linewidth}{!}{
    \begin{tabular}{ccccccccccc}
    \toprule
    \textbf{Datasets} & \textbf{\#Data} & \textbf{Data Percentage} &\textbf{B-1} & \textbf{B-2} & \textbf{B-3} & \textbf{B-4} & \textbf{R-1} & \textbf{R-2}  \\
    \midrule
    Full Dataset & 40,250 & 100\% & 21.23 & 10.40 & 5.98  & 3.84  & 19.49  & 5.58   \\
    Threshold-4 & 21,034 & 52\% &21.34 & 10.63 & 6.15  & 3.93  & 19.56 & 5.58 \\
    Threshold-5 & 23,862 & 59\% & 22.34 & 11.25 & 6.58  & 4.21  & 19.82 & 5.79 \\
    \textbf{Threshold-6} & 24,776 & 62\% &\textbf{23.02} & \textbf{11.84} & \textbf{7.02} & \textbf{4.58} & \textbf{20.2} & \textbf{6.08}  \\
    \bottomrule
    \end{tabular}%
    }
  \label{tab:threshold}%
\end{table*}%

\begin{figure*}
\centering 
\includegraphics[width=1.0\textwidth]{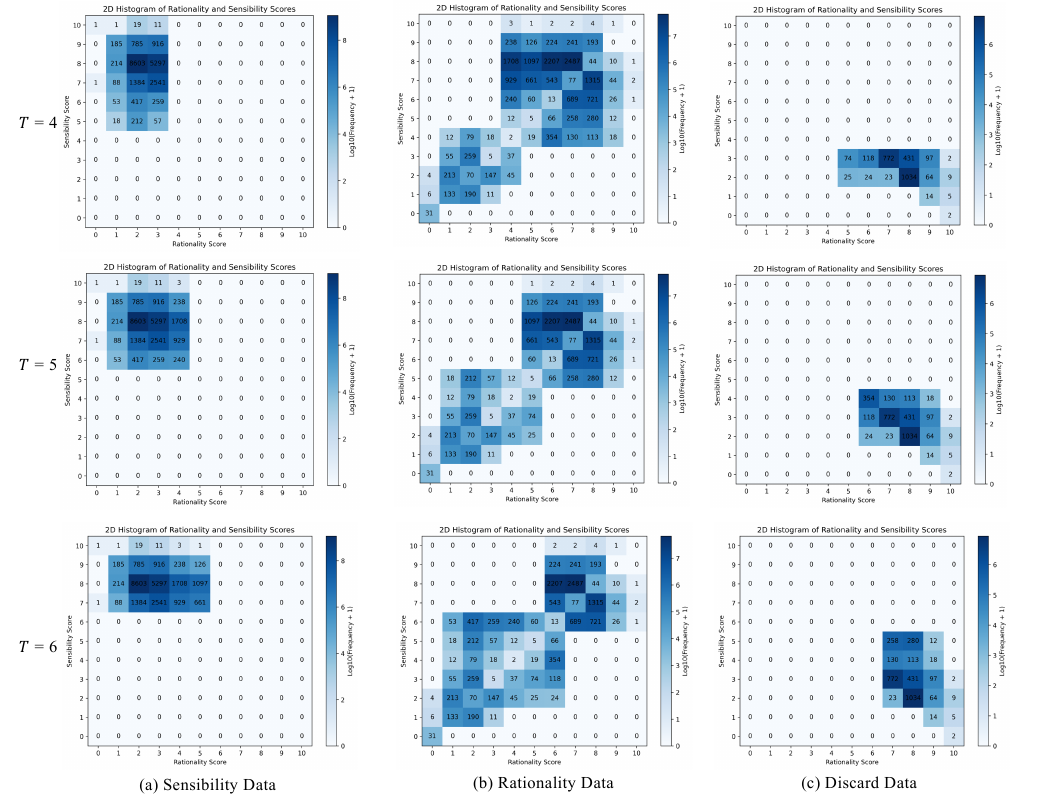} 
\caption{Sensibility and Rationality distribution of different selection thresholds. From top to bottom, respectively. Each row represents a threshold, and from left to right, the columns correspond to sensibility data, rationality data, and discard data.}
\label{Fig.threshold}
\end{figure*}

\subsection{Efficient Sensibility Data}
\label{sec:effect of sen}
To address \textbf{Q2}, in this section, we evaluate the effect of sensibility on empathetic performance. We selected 23,862 (59\%) dialogues from a total of 40,250 based on their sensibility scores. Specifically, dialogues with rationality scores below a certain threshold and sensibility scores above it were included in the sensibility dataset. We then used both the selected sensibility data (Sensibility) and the full dataset (Full F.T.) to fine-tune three commonly used Large Language Models (LLMs): LLaMA3-8B-Instruct, Qwen1.5-7B-Chat, and Qwen1.5-1.8B-Chat.

From Table \ref{tab:sensibility_effect}, we observe that the models trained on the sensibility data outperform those trained on the full dataset across all three models. This demonstrates the efficiency and effectiveness of our data selection method, indicating that focusing on dialogues with higher sensibility and lower rationality improves empathetic performance.

Our results verify the hypothesis that robust sensibility is essential for empathetic performance. Training models with a focused dataset of high sensibility and low rationality dialogues (59\% of the full data) not only reduces the amount of training data required, thereby reducing training costs, but also surpasses all previous SoTA methods. This is evidenced by the consistently higher BLEU (B-1 to B-4), ROUGE (R-1, R-2), and Distinct (Dist-1, Dist-2) scores across all evaluated models. The improvement in these metrics highlights the importance of sensibility in enhancing the empathetic capabilities of LLMs.

In summary, our data selection process, which emphasizes dialogues with high sensibility and low rationality, proves to be an effective strategy for improving empathetic response performance in LLMs. This approach not only enhances model efficiency by reducing the training data size but also leads to SoTA empathetic interactions, as reflected in the superior evaluation metrics.
\subsection{Robustness of Efficient-Empathy}

To address \textbf{Q3}, in this section, we first visualize the data distribution of sensibility, and rationality, and discard data for different thresholds and then explore the effectiveness of different thresholds to demonstrate the robustness of our data selection algorithm. In Figure \ref{Fig.threshold}, we observe the following patterns for thresholds 4, 5, and 6, from top to bottom, respectively. Each row represents a threshold, and from left to right, the columns correspond to sensibility data, rationality data, and discard data. 

It is evident that \textbf{Sensibility Data} are mainly distributed in the upper left corner, indicating high sensibility and low rationality. This distribution suggests that these data points are primarily driven by emotional aspects. \textbf{Rationality Data:} These are predominantly found in the upper right corner, indicating high sensibility and high rationality. This indicates that these data points balance both emotional and logical aspects, making them suitable for empathetic responses. \textbf{Discard Data:} These are mainly distributed in the lower right corner, indicating low sensibility and high rationality. Such data points are not suitable for empathetic responses and are thus categorized as discard data.

The visualized distribution highlights the importance of both sensibility and rationality in determining the suitability of data for empathetic modeling. 

With the selected data based on three different thresholds: 4, 5, and 6, we subsequently train an LLaMA3-8B-Instruct model based on the sensibility data and full dataset. The results are compared with the baseline model which is trained on the full dataset. Table \ref{tab:threshold} shows the performance of our model across different thresholds.

From Table \ref{tab:threshold}, we can see that with hyperparameters set to 4, 5, and 6, and using just above 50\% of the data, our model consistently outperforms the baseline model. This demonstrates the robustness of our data selection algorithm. Specifically, the models trained with Threshold-5 and Threshold-6 show significant improvements across various metrics, including BLEU (B-1 to B-4), ROUGE (R-1, R-2), and Distinct (Dist-1, Dist-2) scores.

As shown in the table, the model trained with Threshold-6, which uses 24,776 training instances, achieves the highest scores in BLEU and ROUGE metrics, indicating better performance in generating diverse and accurate empathetic responses. Notably, the BLEU-1 (B-1) score increased from 21.1 in the full dataset to 23.02 with Threshold-6, and similar trends are observed in BLEU-2 (B-2), BLEU-3 (B-3), BLEU-4 (B-4), ROUGE-1 (R-1), and ROUGE-2 (R-2) scores.

These results validate that our data selection algorithm is robust, effective, and efficient, as it maintains and even enhances model performance with different threshold settings. By selectively training on high-quality data, our approach not only reduces the amount of data required but also improves the overall performance of the LLMs in empathetic response tasks. This demonstrates the stability and effectiveness of our data selection method across varying thresholds, ensuring that the models are trained on the most relevant and impactful data.
\subsection{Human Evaluation}

\begin{figure}
\centering 
\includegraphics[width=8cm]{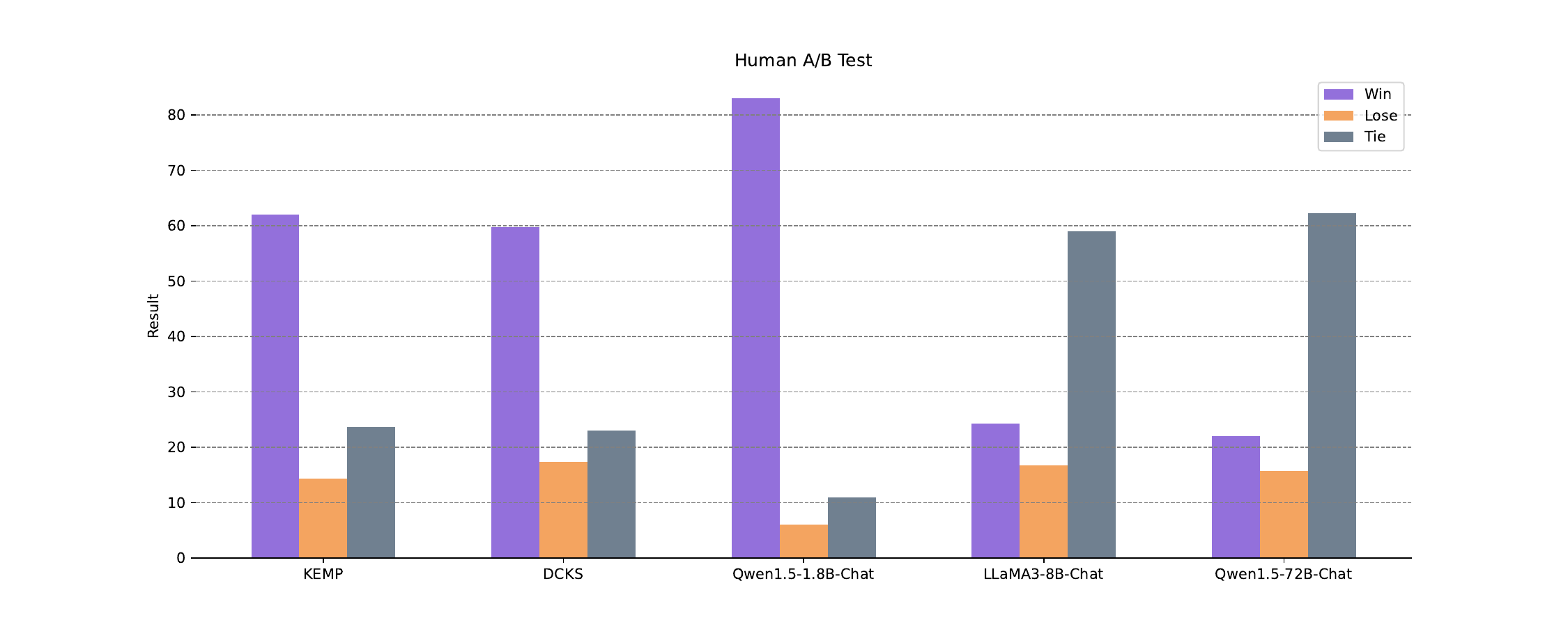} 
\caption{Human A/B test between our MoE model and baseline models.}
\label{Fig.human-ab}
\end{figure}

To address \textbf{Q4}, we conduct the human evaluation A/B testing. The results in Figure~\ref{Fig.human-ab} demonstrate that MoE model(trained on section~\ref{sec:main_exp}) consistently outperforms the baseline models across the dimensions of Coherence, Empathy, Information, and Continuity.

For KEMP and DCKS, which are based on the standard language model BART, our model has a winning rate of around 60\% and a losing rate of approximately 15\%. Notably, the win rate for our model surged to 83.0\% when compared to the 1.8B scale LLM, Qwen1.5-1.8B-Chat. However, as the scale of baseline parameters increases, their effectiveness improves. When compared with Qwen1.5-72B-Chat, the high tie rate of 62.3\% indicates that both models frequently produced comparably effective responses. Nonetheless, our model has a win rate of 22.0\%, higher than the loss rate of 15.7\%. Similarly, the results show a close contest with our model winning 24.3\% and losing 16.7\% in comparison to LLaMA3-8B-Instruct.

\subsection{Case Study}

To address \textbf{Q5}, in this section, we utilize case studies to conduct an in-depth analysis of the impact of our method. We compare the performance of several models, including LLaMA3-8B-Instruct, Qwen1.5-72B-Chat, Qwen1.5-1.8B-Chat, and Full F.T. model(trained on section~\ref{sec:effect of sen}), as well as the Rationality(trained on ~\ref{sec:main_exp}), Sensibility(trained on ~\ref{sec:main_exp}), and MoE experts(trained on ~\ref{sec:main_exp}).

\begin{table}[htbp]
  \centering
  \caption{Case study-1 and the responses from Sensibility, Rationality, MoE, and other models. \textcolor{blue}{Blue text} indicates the presence of rationality information and \textcolor{orange}{Orange text} denotes sensibility expression.}
  \resizebox{\linewidth}{!}{
    \begin{tabular}{m{4.415em}m{4.665em}|m{26.em}}
    \toprule
    \multicolumn{2}{c}{Context} & \textbf{Speaker:} I feel nostalgic when playing old school games with my bro. \\
    \midrule
    \multirow{6}[2]{=}{\vfill Response} &  \multicolumn{1}{l}{LLaMA3-8B-Instruct} & Aww, that's sweet! It's like you're reliving good times with your bro. Nostalgia can be a powerful way to connect with happy memories, doesn't it? \\ 

        \cmidrule{2-3}   & \multicolumn{1}{l}{Qwen1.5-72B-Chat} & That's awesome! Those memories with your bro are precious. Keep enjoying those gaming sessions, they're a great way to bond. \\
        \cmidrule{2-3}   & \multicolumn{1}{l}{Qwen1.5-1.8B-Chat} & Hey bro, I remember when we used to play those old school games back in the day. It's like a flashback to our childhood, and I miss those days so much. Playing games together was such a joy, and we had so much fun. Do you still play those games with your friends? If so, let's catch up soon and reminisce about those memories. \\
       \cmidrule{2-3}   & \multicolumn{1}{l}{Full F.T.} & That is awesome! I am glad you have a good relationship with your brother. \\
       \cmidrule{2-3}   & \multicolumn{1}{l}{Rationality} & \textcolor{blue}{What games do you play?} \\
       \cmidrule{2-3}   & \multicolumn{1}{l}{Sensibility} & \textcolor{orange}{I can imagine.} It is always fun to play old school games with your siblings. \\
       \cmidrule{2-3}   & \multicolumn{1}{l}{MoE Model} & \textcolor{orange}{I remember those games too.} \textcolor{blue}{What kind of games did you play?} \\
    \midrule
    \multicolumn{2}{c}{Golden Respose} & Nice , me too , are you playing nintendo games ? \\
    \bottomrule
    \end{tabular}%
    }
  \label{tab:case-1}%
\end{table}%


As shown in Table~\ref{tab:case-1}, LLaMA3-8B-Instruct, while providing warm and positive responses, tends to lack specific resonant experiences. This results in replies that feel somewhat generic and superficial, failing to delve deeply into the speaker's unique context and emotions. Similarly, Qwen1.5-72B-Chat's responses, though affirming and encouraging, also suffer from a lack of specific resonant experiences and interactive guidance. In contrast, the Rationality expert's responses lack emotional recognition, failing to mirror the speaker's sentiments. Besides, although both the Sensibility expert and Baseline model demonstrate understanding towards the speaker, their responses lack the inclination to continue the conversation.

Our approach aims to address these shortcomings by combining sensibility and rationality to foster a deeper connection. By immediately acknowledging that the listener has had similar experiences in the first sentence of our response, we quickly establishes empathy and demonstrates a clear understanding of the speaker's feelings. Subsequently, the second sentence asks specific questions about the speaker's experience, which not only shows a genuine interest in their story but also actively encourages the conversation to continue, making the interaction more personalized and dynamic.

In Table~\ref{tab:case-2}, the response of LLaMA3-8B-Instruct, "\textit{You're not usually anxious about trains}" assumes the speaker's typical state of mind regarding trains. By making this assumption, LLaMA risks misrepresenting or misunderstanding the speaker's feelings or experiences. If the speaker does not identify with this assumption, they may feel misunderstood or unfairly labeled. Similarly, Qwen1.5-72B-Chat's approach fails to adequately address the speaker's unique situation. This generality can lead to a lack of engagement and depth in the dialogue.

In analyzing different expert approaches, we distinguish between the Rationality expert and the Sensibility expert. The Rationality expert primarily focuses on identifying and understanding the causes of the horror of the train. While it may offer logical and practical solutions, it often fails to consider the speaker's emotional state. Conversely, the Sensibility expert emphasizes understanding and empathizing with the user's sacred feeling. However, it may not effectively guide the conversation towards practical solutions or a deeper understanding of the issue at hand.

Our proposed approach seeks to balance these two expert perspectives, aiming to both understand the user's state and address the underlying issues without making unwarranted assumptions. By showing genuine concern for whether the user is experiencing difficulties through "\textit{Did you have trouble}", we create a supportive environment that validates the user's feelings and experiences. Furthermore, instead of making assumptions, we directly ask about the cause of the problem through "\textit{Why?}". This method allows the speaker to provide their own explanation, ensuring that their perspective is accurately represented.

\begin{table}[htbp]
  \centering
  \caption{Case study-2 and the responses from Sensibility, Rationality, MoE, and other models. \textcolor{blue}{Blue text} indicates the presence of rationality information and \textcolor{orange}{Orange text} denotes the sensibility expression.}
  \resizebox{\linewidth}{!}{
    \begin{tabular}{m{4.415em}m{4.665em}|m{26.em}}
    \toprule
    \multicolumn{2}{c}{Context} & \textbf{Speaker:} I was so scared of going onto the train ! \\
    \midrule
    \multirow{6}[2]{=}{ \vfill Response} &  \multicolumn{1}{l}{LLaMA3-8B-Instruct} & Ah, girl, what happened? You're not usually anxious about trains. Did something specific happen that triggered this fear? \\ 

        \cmidrule{2-3}   & \multicolumn{1}{l}{Qwen1.5-72B-Chat} & I get that, trains can be intimidating, especially if you're not used to them. But you made it on, that's brave! Next time will be easier. \\

        \cmidrule{2-3}   & \multicolumn{1}{l}{Qwen1.5-1.B-Chat} & Oh, I'm sorry to hear that you're feeling scared about going on the train. It's completely normal to feel anxious about taking a big step like this, especially if it's your first time. Remember that taking a train can be a relatively safe and familiar experience, especially if you've taken the train before. Here are some tips that might help you feel more comfortable: 1. Plan ahead: Make sure you have a good understanding of the train schedule and route. Familiarize yourself with the train's schedule, the stations, and the times when the train arrives and departs. Knowing the journey ahead can help alleviate some anxiety \\
    
    \cmidrule{2-3} &
    \multicolumn{1}{l}{Full F.T.} & Oh no! What happened? \\
    \cmidrule{2-3}
          & \multicolumn{1}{l}{Rationality} & \textcolor{blue}{What was wrong} with the train? \\
       \cmidrule{2-3}   & \multicolumn{1}{l}{Sensibility} & I can imagine! I have been on trains before and \textcolor{orange}{i was scared} too! \\
       \cmidrule{2-3}   & \multicolumn{1}{l}{MoE Model} & \textcolor{blue}{Why?} \textcolor{orange}{Did you have trouble} getting on? \\
    \midrule
    \multicolumn{2}{c}{Golden Respose} & Was it your first time riding on one ? Is there something in particular that makes you afraid ? \\
    \bottomrule
    \end{tabular}%
    }
  \label{tab:case-2}%
\end{table}%

In Table~\ref{tab:case-3}, we analyze the approaches of LLaMA3-8B-Instruct and Qwen1.5-72B-Chat, followed by a description of our proposed method.

LLaMA3-8B-Instruct expresses concern about the movie's suitability for the speaker's daughter but does so in a manner that can be perceived as questioning and critical, potentially making the speaker feel judged or unsupported. This can lead to feelings of misunderstanding and frustration, negatively impacting the flow and depth of the conversation. Similarly, Qwen1.5-72B-Chat's response, while less overtly critical than LLaMA's, still carries negative implications. By implicitly questioning the suitability of the movie choice for her daughter, Qwen1.5-72B-Chat may inadvertently suggest that the speaker's judgment is flawed, causing the speaker to feel defensive and less confident in their decisions.

Our proposed method focuses on respectful and open-ended inquiry, avoiding assumptions and fostering supportive dialogue. By directly asking about the daughter's reaction to the movie without making any assumptions, our approach respects the speaker's judgment. Open-ended questions encourage the speaker to share more details about their experience.

In conclusion, our approach excels by maintaining simplicity and directness, avoiding assumptions, and respecting the speaker's judgment. This method not only encourages the speaker to share more freely but also enhances the overall quality and depth of the conversation.
\begin{table*}[htbp]
  \centering
  \caption{Rationality experts are further integrated with sensibility experts into an MoE model. When the rationality expert is replaced with the base model and the discard expert, a performance drop is observed.}
    \resizebox{0.8\linewidth}{!}{
    \begin{tabular}{clcccccc}
    \toprule
          \textbf{Base Model}& \textbf{Models} & \textbf{B-1} & \textbf{B-2} & \textbf{B-3} & \textbf{B-4} & \textbf{R-1} & \textbf{R-2} \\
    \midrule
          \multirow{4}[1]{*}{LLaMA3-8B-Instruct} 
          & Sensibility & 22.34 & 11.25 & 6.58  & 4.21  & 19.82 & 5.79  \\
          & Rationality & 21.4  & 10.47 & 6.08  & 3.88  & 19.46 & 6.18  \\
          & MoE Model   &  \textbf{23.04} & \textbf{11.62} & \textbf{6.68} & \textbf{4.22} & \textbf{20.28} & \textbf{6.15} \\
          \rowcolor{Violet!15} 
           & \textit{+ increase } & $\blacktriangle$ 0.70   & $\blacktriangle$ 0.37  & $\blacktriangle$ 0.10  & $\blacktriangle$ 0.01  & $\blacktriangle$ 0.46  & $\blacktriangle$ 0.36   \\
           
          & struc-(a) & 22.16 & 11.08 & 6.33  & 3.98   & 20.12 & 5.92  \\
          \rowcolor{Violet!15} 
            &\textit{- decrease } & $\blacktriangledown$ 0.88   & $\blacktriangledown$ 0.54  & $\blacktriangledown$ 0.35  & $\blacktriangledown$ 0.24  & $\blacktriangledown$ 0.16  & $\blacktriangledown$ 0.23    \\
            
          & struc-(b) & 21.30 & 10.75  & 6.23  & 3.97   & 20.13 & 6.13  \\
          \rowcolor{Violet!15} 
            &\textit{- decrease } & $\blacktriangledown$ 1.74   & $\blacktriangledown$ 0.87  & $\blacktriangledown$ 0.45  & $\blacktriangledown$ 0.25  & $\blacktriangledown$ 0.15  & $\blacktriangledown$ 0.02    \\

        & struc-(c)   & 22.41 & 11.33  & 6.55  & 4.16   & 20.21 & 6.14  \\
            \rowcolor{Violet!15} 
            &\textit{- decrease } & $\blacktriangledown$ 0.63   & $\blacktriangledown$ 0.29  & $\blacktriangledown$ 0.13  & $\blacktriangledown$ 0.06  & $\blacktriangledown$ -  & $\blacktriangledown$ 0.01  \\
          
          & struc-(d)   & 22.39 & 11.18  & 6.39  & 4.04   & 19.87 & 5.88  \\
            \rowcolor{Violet!15} 
            &\textit{- decrease } & $\blacktriangledown$ 0.65   & $\blacktriangledown$ 0.44  & $\blacktriangledown$ 0.29  & $\blacktriangledown$ 0.18  & $\blacktriangledown$ 0.41  & $\blacktriangledown$ 0.27    \\
            
    \midrule
        \multirow{4}[1]{*}{Qwen1.5-1.8B-Chat} 
          & Sensibility & 18.2  & 7.5   & 3.73  & 2.16  & 15.74 & 3.48  \\
          & Rationality & 17.62 & 6.91  & 3.31  & 1.83  & 14.81 & 3.06  \\
          
          & MoE Model & \textbf{22.01} & \textbf{11.16}  & \textbf{6.53}  & \textbf{4.21}   & \textbf{20.27} & \textbf{6.30}  \\
            \rowcolor{Violet!15} 
           & \textit{+ increase } & $\blacktriangle$ 3.81   & $\blacktriangle$ 3.66  & $\blacktriangle$ 2.80  & $\blacktriangle$ 2.05  & $\blacktriangle$ 4.53  & $\blacktriangle$ 2.82   \\
          
          
    \bottomrule
    \end{tabular}%
    }
  \label{tab:moe_effect}%
\end{table*}%

\subsection{Ablation Study}
\begin{table}[htbp]
  \centering
  \caption{Case study-3 and the responses from Sensibility, Rationality, MoE, and other models. \textcolor{blue}{Blue text} indicates the presence of rationality information and \textcolor{orange}{Orange text} denotes the sensibility expression.}
  \resizebox{\linewidth}{!}{
    \begin{tabular}{m{4.415em}m{4.665em}|m{26.em}}
    \toprule
    \multicolumn{2}{c}{Context} & \textbf{Speaker:} I introduced my oldest daughter to the movie fight club the other day . \\
    \midrule
    \multirow{6}[2]{=}{ \vfill Response} & \multicolumn{1}{l}{LLaMA3-8B-Instruct} & Whoa, what were you thinking?! Fight Club is definitely not suitable for kids. How did that go? \\ 

        \cmidrule{2-3}   & \multicolumn{1}{l}{Qwen1.5-72B-Chat} & Oh cool, how did she like it? Was it too intense for her or did she enjoy the twist? \\

        \cmidrule{2-3}   & \multicolumn{1}{l}{Qwen1.5-1.8B-Chat} & That sounds like a fun and exciting experience for your daughter! What did your daughter enjoy about the movie and what did she think about the themes and messages it conveyed? Did she have any questions or concerns about the movie or the characters? It's always great to hear how our favorite movies can resonate with others and provide a meaningful perspective. \\

        \cmidrule{2-3}   & \multicolumn{1}{l}{Full F.T.} & Oh wow, that is a very mature movie. I hope she was prepared for it. \\
        \cmidrule{2-3}  & \multicolumn{1}{l}{Rationality} & I am not sure that is a good idea. The movie is rated r and it is \textcolor{blue}{not suitable} for children. \\
        \cmidrule{2-3}  & \multicolumn{1}{l}{Sensibility} & Oh wow, that is \textcolor{orange}{a pretty intense} movie. I hope she was not too disturbed by it. \\
        \cmidrule{2-3}  & \multicolumn{1}{l}{MoE Model} & That is a \textcolor{orange}{good movie}. \textcolor{blue}{Did she like it?} \\
    \midrule
    \multicolumn{2}{c}{Golden Respose} & Interesting choice . Did she like it ? \\
    \bottomrule
    \end{tabular}%
    }
  \label{tab:case-3}%
\end{table}%

\begin{figure}
\centering
\includegraphics[width=0.5\textwidth]{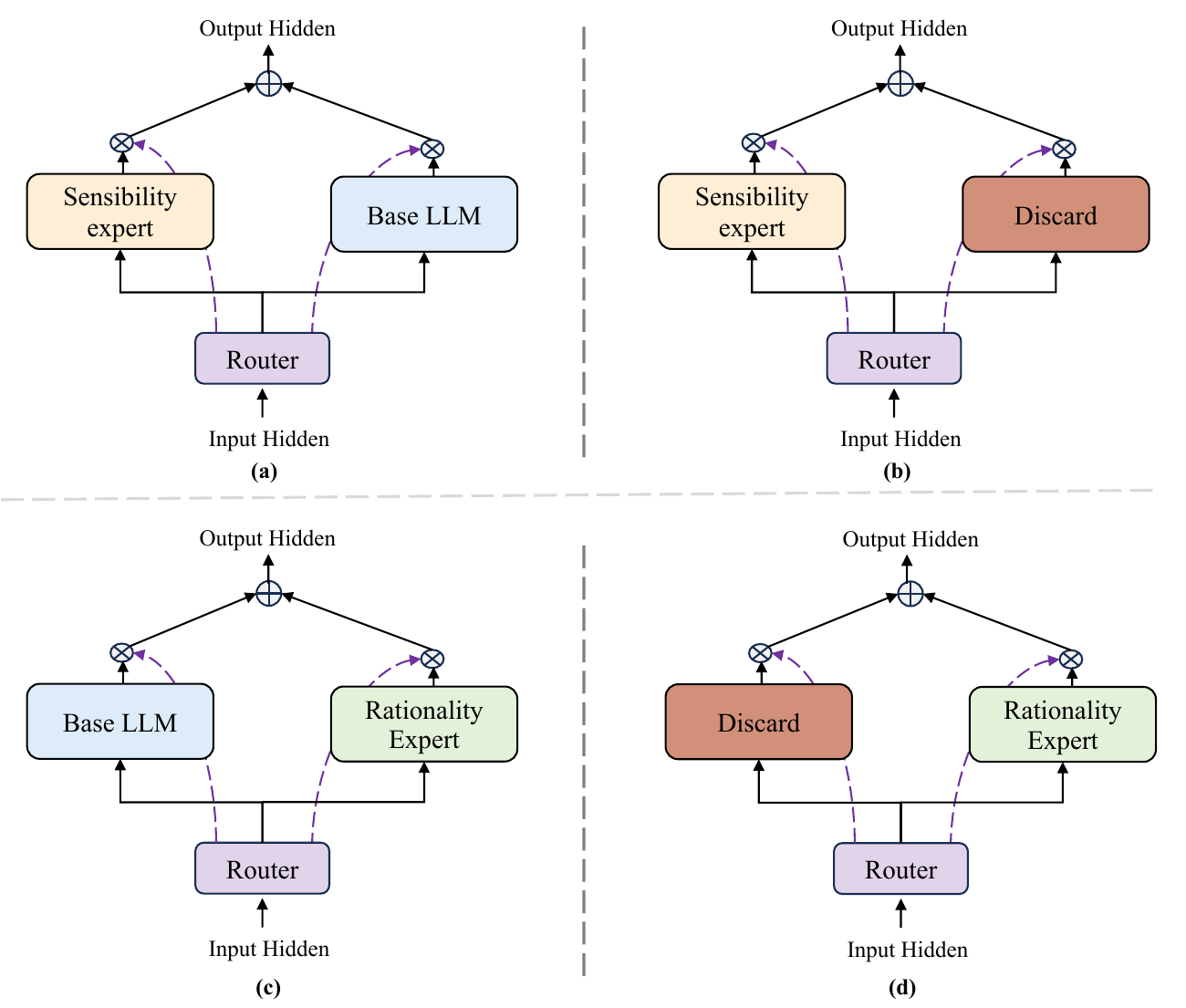}
\caption{The overall ablation MoE model structure.}
\label{Fig.model-moe-3}
\end{figure}
To address \textbf{Q6}, following section \ref{section4}, we first use the selected datasets to train the sensibility, rationality, and discard experts based on the LLaMA3-8B-Instruct and Qwen1.5-1.8B-Chat models. Then, we use the sensibility and rationality experts to train an MoE model based on LLaMA3-8B-Instruct and Qwen1.5-1.8B-Chat. From Table \ref{tab:moe_effect}, we can see that both models outperform a single expert and achieve SoTA performance. However, it is unclear whether the SoTA performance is due to the MoE structure or the selected datasets. To evaluate the effects of the selected datasets and the MoE structure, we conduct the following ablation experiments:

\textbf{Replace the rationality expert}: We replace the rationality expert to demonstrate its effectiveness. First, we replace the rationality expert with the base model, as shown in Figure \ref{Fig.model-moe-3}(a). Then, we replace the rationality expert with the discard model, as shown in Figure \ref{Fig.model-moe-3}(b).

\textbf{Replace the sensibility expert}: We replace the sensibility expert to demonstrate its effectiveness. First, we replace the sensibility expert with the base model, as shown in Figure \ref{Fig.model-moe-3}(c). Then, we replace the sensibility expert with the discard model, as shown in Figure \ref{Fig.model-moe-3}(d).

From Table \ref{tab:moe_effect}, we can see that modifying either the rationality expert or the sensibility expert significantly reduces the performance of both the LLaMA3-8B-Instruct and Qwen1.5-1.8B-Chat based MoE models.

\FloatBarrier 
\section{Conclusion}
Empathy is a critical component of human social interaction and communication. However, high-quality empathetic data remains scarce, necessitating efficient and effective data selection algorithms. This paper presents Efficient-Empathy, a new approach to empathetic data selection by leveraging LLM-rated sensibility and rationality scores. Remarkably, our model outperforms all previous models using only 59\% of the selected sensibility data. Furthermore, by integrating the selected sensibility and rationality data with a MoE model, it achieves even higher SoTA performance without adding extra data.







\newpage

\bibliographystyle{ACM-Reference-Format}
\bibliography{main}

\end{document}